\documentclass[10pt,twocolumn,letterpaper]{article}

\usepackage{cvpr}              
\definecolor{cvprblue}{rgb}{0.21,0.49,0.74}
\usepackage[pagebackref,breaklinks,colorlinks,allcolors=cvprblue]{hyperref}
\usepackage{graphicx} 
\usepackage{caption}
\usepackage{tabularx,booktabs,multirow,array}

\def\paperID{21305} 
\def\confName{CVPR}
\def\confYear{2026}

\title{Polyphony: Diffusion-based Dual-Hand Action Segmentation with Alternating Vision Transformer and Semantic Conditioning}

\author{
Hao Zheng$^{1}$ \quad
Hu Wang$^{2}$ \quad
Tiantian Zheng$^{1}$ \quad
Prajjwal Bhattarai$^{1}$ \quad
Tuka Alhanai$^{1}$\\[0.5em]
$^{1}$New York University Abu Dhabi, Abu Dhabi, United Arab Emirates\\
$^{2}$Mohamed bin Zayed University of Artificial Intelligence, Abu Dhabi, United Arab Emirates\\
{\tt\small \{h.zheng, tz545, pb2276, tuka.alhanai\}@nyu.edu \quad hu.wang@mbzuai.ac.ae}
}
\begin{document}
\maketitle

\begin{abstract}
Dual-hand action segmentation, densely predicting actions for both hands from untrimmed videos, is essential for understanding complex bimanual activities. However, it poses several unique challenges: complex inter-hand dependencies, visual asymmetry between hands, representation conflicts where the dominant hand monopolizes gradients, and semantic ambiguity in fine-grained actions. We propose Polyphony, a three-stage method to address these challenges through: (1) an Alternating Dual-Hand Vision Transformer that alternates training between left- and right-hand mini-batches to ensure balanced gradient contributions from both hands while sharing a spatio-temporal encoder; (2) Semantic Feature Conditioning that aligns visual features with structured, compositional action descriptions to enhance discrimination of semantically similar actions; and (3) Diffusion-Based Segmentation with cross-hand feature fusion for inter-hand coordination and adaptive loss weighting for balancing performance. Polyphony achieves state-of-the-art on both dual-hand datasets (HA-ViD, ATTACH) with improvements up to 16.8 points, and on the single-stream Breakfast dataset (82.5\%), outperforming the prior best method that uses a 12× larger backbone. Notably, our unified model with a single shared backbone surpasses baselines requiring separate per-hand models. Code is at \url{https://github.com/x-labs-xyz/Polyphony-Dual-hand-Action-Segmentation}.
\end{abstract}

\section{Introduction}

Understanding human activities from video is a cornerstone of computer vision, with applications ranging from robotics and human-computer interaction to video content analysis \cite{ZHENG2025102976,pareek_survey_2021,sharma_aligning_2025}. While significant progress has been made in action recognition and temporal action segmentation \cite{videomaev2,ms-tcn,diffact,FACT}, the vast majority of methods model human activity as a single stream of actions. This paradigm falls short in capturing the intricate dynamics of bimanual activities, which are ubiquitous in real-world tasks such as assembly, cooking, and surgery \cite{ha-vid,9720489,yilmaz_continuous_2022}. 

Dense, frame-wise segmentation of dual-hand actions presents a set of unique and critical challenges that are not addressed by conventional methods: (1) \textbf{Complex Inter-Hand Dependencies}: The two hands exhibit complex temporal relationships, switching between coordinated, cooperative, and independent action patterns; (2) \textbf{Visual Asymmetry}: The same action can manifest with distinct appearance and motion patterns for the left versus right hand; (3) \textbf{Representation Conflict}: In a joint model, the dominant hand may dominate gradient updates, leading to suboptimal feature learning for the non-dominant hand; (4) \textbf{Semantic Ambiguity}: Relying solely on visual features can fail to capture fine-grained semantic distinctions between similar actions (e.g., ``screw nut onto bolt" vs. ``screw nut onto shaft").

To tackle these challenges, we propose Polyphony, a three-stage method for dual-hand action segmentation. Inspired by polyphonic music, where independent melodies harmonize simultaneously, Polyphony models dual-hand actions as interdependent streams requiring both individual understanding and collective coordination. To address representation conflict and visual asymmetry, we introduce an Alternating Dual-Hand Vision Transformer (ADH-ViT) that alternates training between left- and right-hand mini-batches, preventing gradient dominance while maintaining a shared encoder for efficient learning. To combat semantic ambiguity, we develop Semantic Feature Conditioning that enriches visual features by aligning them with structured, compositional language descriptions (e.g., verb-object-tool decomposition). To model complex inter-hand dependencies, we design a Diffusion-Based Segmentation model with cross-hand feature fusion for coherent dual-hand predictions and adaptive loss weighting for balanced training. The shared encoder design reflects the holistic perception of bimanual activities by humans \cite{swinnen_two_2004}, while diffusion-based iterative refinement mirrors humans' progressive action understanding through repeated predictions and corrections \cite{kilner_predictive_2007}. Polyphony’s modular and versatile design (i) permits independent component improvement and (ii) allows ADH-ViT to operate as a standalone action-recognition model for broader downstream tasks.

Extensive experiments on three challenging benchmarks validate our approach. On the dual-hand datasets HA-ViD and ATTACH, our method sets a new state-of-the-art, significantly outperforming previous work. Our model additionally outperforms others on the Breakfast dataset for single-stream action segmentation, demonstrating generalizability beyond the bimanual domain. Comprehensive ablation studies confirm the efficacy of each core component, while qualitative analyses highlight Polyphony's capacity to address the core challenges of complex inter-hand dependencies, visual asymmetry, and semantic ambiguity in dual-hand action segmentation.

\section{Related Work}

\textbf{Temporal Action Segmentation.} Temporal action segmentation (TAS) aims to predict frame-wise action labels in untrimmed videos, a fundamental task for understanding procedural activities. Early approaches established Temporal Convolutional Network (TCN) as the foundation for modeling long-range temporal dependencies \cite{tcn}. Building on this foundation, subsequent work introduced multi-stage refinement strategies \cite{ms-tcn}, boundary-aware losses \cite{BCN}, and graph-based modeling \cite{9156462} to address persistent over-segmentation errors. The field evolved further with transformer-based architectures \cite{ASformer, FACT}, which leverage self-attention mechanisms to capture long-range dependencies more effectively. Most recently, diffusion models have emerged as a powerful paradigm, formulating action segmentation as conditional generation through iterative denoising processes \cite{diffact}. However, existing TAS methods predominantly focus on single-stream action sequences and do not address the unique challenges of dual-hand scenarios, where two interdependent action streams must be modeled simultaneously while attending to their complex coordination patterns. Zheng et al. proposed two dual-hand action segmentation methods, DuHa \cite{DuHa} and DuCAS \cite{DuCAS}, specifically for bimanual assembly scenarios, but both methods require ground-truth object bounding boxes as input, limiting their general applicability. In this paper, we present a diffusion-based dual-hand action segmentation model that simultaneously predicts actions for both hands without requiring object annotations.

\textbf{Vision-Language Alignment and Semantic Conditioning.} Integrating semantic knowledge with visual understanding has shown significant benefits across various vision tasks \cite{10445007}. Vision-language models such as CLIP \cite{clip} and ALIGN \cite{align} align visual and textual embeddings through contrastive learning, enabling zero-shot transfer. These models have been successfully adapted to video understanding through temporal attention \cite{X-CLIP} and video-specific prompting \cite{ActionCLIP}, achieving strong performance on action recognition. In temporal action segmentation, FACT \cite{FACT}  leverages video transcripts, parsed from frame-wise action labels, to construct action-level tokens that can enhance frame-wise action predictions through cross-attention. However, relying on simple action labels limits the semantic richness needed to disambiguate actions with fine-grained semantic distinctions. Compositional action understanding \cite{something-else} addresses this by explicitly modeling verb-object structure, with recent approaches exploiting knowledge bases \cite{dark}, semantic disentanglement \cite{deformer}, and grammar-based decomposition \cite{kcmm} for distinguishing semantically nuanced actions. We extend this paradigm by incorporating structured compositional descriptions into our dual-hand segmentation framework, decomposing frame-wise action labels into verb, object, and tool components \cite{hr-sat}. A TCN is applied to align visual features with these language-model–encoded descriptions, enabling semantic conditioning that addresses the semantic ambiguity.

\section{Methodology}

\subsection{Overview and Problem Formulation}

\begin{figure}
    \centering
    \includegraphics[width=1\linewidth]{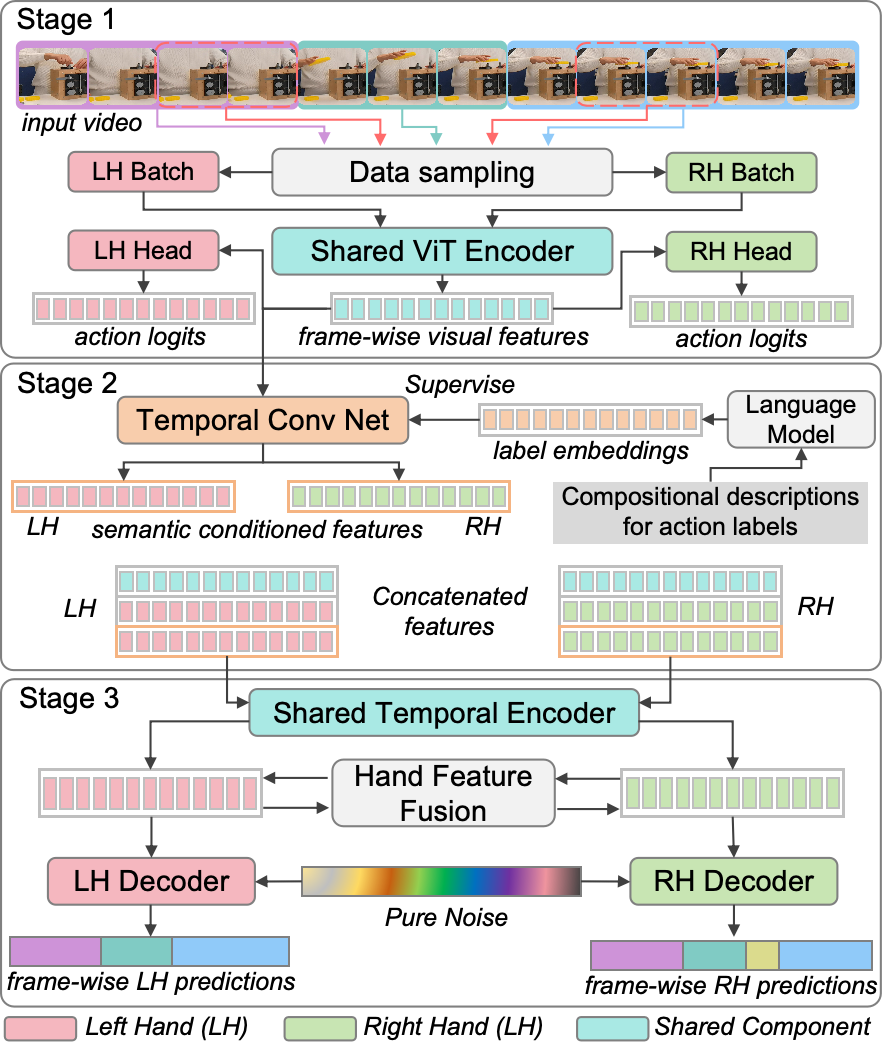}
    \caption{Overview of Polyphony, a three-stage dual-hand action segmentation method. Stage 1 extracts dual-hand features via a shared ViT and hand-specific classification heads; Stage 2 performs semantic feature conditioning by aligning visual features with compositional action descriptions; Stage 3 conducts diffusion-based segmentation with cross-hand feature fusion. The modular architecture enables: (1) flexible deployment—handling both dual-hand and single-stream tasks with potential extension to multi-agent scenarios; (2) versatile application—the ViT in Stage 1 can operate as a standalone action recognition model; (3) modular design—each component can be improved independently.}
    \label{fig:overall_framework}
\end{figure}

Given an untrimmed video, our goal is to perform dense frame-level action segmentation for both the left hand (LH) and right hand (RH) simultaneously through one model. Let $\mathcal{V} = \{v_\textit{t}\}_{t=1}^T$ denote a video sequence of $T$ frames. We aim to predict frame-wise action label sequences $Y^h = \{y_t^h\}_\textit{t=1}^T$ , $h\in\{LH,RH\}$, where $y_t \in \{1, 2, ..., C\}$ represents one of $C$ action classes. To address the dual-hand action segmentation challenges, we propose Polyphony (Figure \ref{fig:overall_framework}), which has three stages: (1) Dual-Hand Feature Extraction, (2) Semantic Feature Conditioning, and (3) Diffusion-Based Dual-Hand Action Segmentation. 

\subsection{Stage 1: Dual-Hand Feature Extraction}
We propose ADH-ViT to extract hand-specific features for the subsequent action segmentation stage.

\subsubsection{Data Sampling}
To adapt untrimmed videos for clip-based ADH-ViT training, complementary sampling strategies are applied independently to each hand.

\textbf{Segment-Based Sampling.} From frame-wise annotations $\{y_t^h\}_{t=1}^T$, we extract coherent action segments. Formally, for action class $c_i$, the set of segments is defined as:
\begin{equation}
    S_{seg}^h = \left\{(t_{\text{start}}, t_{\text{end}}, c_i) \mid y_t^h = c_i \ \forall t \in [t_{\text{start}}, t_{\text{end}}]\right\}
\end{equation}
Each segment $(t_{\text{start}}, t_{\text{end}}, c_i)$ yields a training clip $x_{seg}$ labeled with action class $c_i$.

\textbf{Random Clip Sampling.} To enhance temporal diversity and augment training data, we randomly sample $n_{clip}$ clips of $l_{clip}$ consecutive frames throughout the video. The label for each random clip $x_{clip}$ is determined by the action class of its middle frame.

The complete samples $\mathcal{X}^h = \{x_{seg}^h\} \cup \{x_{clip}^h\}$. The trimmed video clip dataset for each hand is obtained, $\mathcal{D}_{trim}^h = \{(x_i^h,y_i^h)\}_{i=1}^{N_h}$, where $N_h$ denotes the total number of video clips for hand $h$.

\subsubsection{Alternating Dual-Hand Vision Transformer}
The architecture of ADH-ViT consists of:

\textbf{Shared Spatio-Temporal Encoder.} A ViT \cite{ViT} backbone $\mathcal{E}_\phi$ processes video clips via tubelet embeddings. Given $x\in\mathbb{R}^{3\times L \times H \times W} \ (x \in \mathcal{D}_{trim})$ , it is partitioned into non-overlapping tubelet patches of size $l \times p \times p$.
\begin{equation}
z^h = \mathrm{Tubelet}(x^h; l, p) + \mathbf{E}_{\text{pos}} \in \mathbb{R}^{N \times D}
\end{equation}
where $N=\frac{L}{l} \cdot \frac{H}{p} \cdot \frac{W}{p}$ is the number of patches, $D$ is the embedding dimension, and $\mathbf{E}_{\text{pos}}\in \mathbb{R}^{N \times D}$ are sinusoidal positional embeddings. The $\mathrm{Tubelet}$ embedding is implemented as a 3D convolution. Then, the shared spatio-temporal feature is encoded as:
\begin{equation}
    e^h = \mathcal{E}_\phi(z^h) \in \mathbb{R}^D
\end{equation}
\textbf{Dual Classification Heads.} Two independent linear classifiers map shared spatio-temporal features to hand-specific predictions:
\begin{equation}
\hat{y}^{h} = \text{softmax}(\mathbf{W}_{h} e^{h} + b_{h}) \in \mathbb{R}^{C}, \quad h \in \{LH, RH\}
\end{equation}
where $\textbf{W}_{h} \in \mathbb{R}^{C \times D}$ and $b_{h} \in \mathbb{R}^{C}$ are learnable parameters.

\subsubsection{Alternating Training Strategy}
To ensure balanced multi-task optimization, we alternate between LH and RH mini-batches every $\Delta$ steps. Let $\tau(j)\in \{LH,RH\}$ denote the active task at training step $j$:
\begin{equation}
    \tau(j) = 
\begin{cases} 
\text{LH} & \text{if } \lfloor j/\Delta \rfloor \mod 2 = 0 \\ 
\text{RH} & \text{otherwise}
\end{cases}
\end{equation}
At each step, we sample a mini-batch exclusively from the dataset corresponding to $\tau(j)$:
\begin{equation}
    \mathcal{B}^{(j)} = 
\begin{cases} 
\text{sample}(\mathcal{D}_{trim}^{LH},B) & \text{if } \tau(j)=LH \\ 
\text{sample}(\mathcal{D}_{trim}^{RH},B) & \text{if } \tau(j)=RH
\end{cases}
\end{equation}
where $B$ is the batch size. The loss at step $k$ is computed using only the active head:
\begin{equation}
\mathcal{L}^{(j)} = \frac{1}{B} \sum_{i=1}^{B} \mathcal{L}_{CE}(\hat{y}_{i}^{\tau(j)}, y_{i}^{\tau(j)})
\end{equation}
where $\mathcal{L}_{CE}$ is cross-entropy loss. The gradient update modifies both the shared backbone and the active head. The advantage of alternating training over naive joint training is that it prevents tasks by one hand from dominating the gradients updates.

\subsubsection{Dense Frame-Wise Feature Extraction}
To perform action segmentation on untrimmed videos, we extract dense frame-wise features using the trained ADH-ViT. During feature extraction, we apply a $l_{clip}$-long sliding window with symmetric padding over the entire video $\mathcal{V} = \{v_\textit{t}\}_{t=1}^T$ to get $T$ video clips. Each clip $x_t=\{v_i\}_{i=t-l_{clip}/{2}+1}^{t+l_{clip}/{2}}$ is processed through the trained encoder to extract both shared and hand-specific features:
\begin{equation}
\begin{aligned}
    e_t^{\text{shared}} = \mathcal{E}_{\phi}(\mathrm{Tubelet}(x_t; l, p) + \mathbf{E}_{\text{pos}})
\end{aligned}
\end{equation}
\begin{equation}
\begin{aligned}
    e_t^{h, \text{class}} = \mathbf{W}_h e_t^{\text{shared}}+b_{h}
\end{aligned}
\end{equation}
where the $e_t^{\text{shared}}$ and $e_t^{h, \text{class}}$ are the shared backbone feature and hand-specific classification logit for video clip $x_t$.

\subsection{Stage 2: Semantic Feature Conditioning}
Visual features alone may conflate semantically distinct but visually similar actions. To enhance semantic discrimination, a TCN is applied to align visual features with semantic embeddings.
\subsubsection{Structured Action Description}
We represent each action class $c$ as a structured, compositional semantic description (parsed from the original label), following the methodology of HR-SAT \cite{hr-sat}:
\begin{equation}
\begin{aligned}
 D_c = \text{``Action verb is } av_c; \text{manipulated object is } mo_c; \\ \text{target object is } to_c; \text{tool is } tl_c\text{.''} 
\end{aligned}
\end{equation}
For example, ``screw nut onto bolt" becomes: \textit{``Action verb is screw; manipulated object is nut; target object is bolt; tool is null"}. This structured format provides explicit semantic attributes that distinguish similar actions.

Each description $D_c$ is encoded using a pre-trained language model:
\begin{equation}
    e_c = \text{LM}(D_c) \in \mathbb{R}^{D_{sem}}
\end{equation}
where $D_{sem}$ is the semantic embedding dimension. These embeddings serve as supervision signals for semantic conditioning. These descriptions are used only for training, not inference, preventing label semantics leakage.

\subsubsection{TCN for Semantic Alignment}
We employ a multi-layer TCN to model temporal context and align visual features with semantic embeddings. The TCN consists of $M$ residual blocks with exponentially increasing dilation:
\begin{equation}
    h_m = \text{TCNBlock}(h_{m-1}; d_m = 2^m), m=1,\dots, M
\end{equation}
where $h_0$ is shared backbone feature $e_t^{\text{shared}}$. 
The TCN output is projected into the semantic space:
\begin{equation}
\begin{aligned}
e_t^{h,\text{sem}} &= \text{LN}(\mathbf{W}_{sem} h_M + b_{sem}) \in \mathbb{R}^{D_{\text{sem}}}
\end{aligned}
\end{equation}
where $\mathbf{W}_{sem}$ and $b_{sem}$ are learnable projection matrices.
\subsubsection{Adaptive Alignment Loss}
We train the TCN using an adaptive multi-objective loss that combines cosine similarity and mean squared error:
\begin{equation}
    \mathcal{L}_{\text{align}} = \alpha \mathcal{L}_{\text{cosine}} + (1 - \alpha) \mathcal{L}_{\text{MSE}}
\end{equation}
where:
\begin{equation}
    \mathcal{L}_{\text{cosine}} = 1 - \frac{e_t^{h, \text{sem}} \cdot e_{y_t}}{\|e_t^{h, \text{sem}}\|_2 \|e_{y_t}\|_2}
\end{equation}
\begin{equation}
    \mathcal{L}_{\text{MSE}} = \|e_t^{h, \text{sem}} - e_{y_t}\|_2^2
\end{equation}
The weighting factor $\alpha$ balances directional alignment (cosine loss) with magnitude matching (MSE loss).
\subsubsection{Motion-Action-Semantic Feature Concatenation}
To provide comprehensive representations for fine-grained action segmentation, we construct a Motion-Action-Semantic (MAS) feature by concatenating three complementary feature types for each hand:
\begin{equation}
    E_t^h=[e_t^{shared}, e_t^{h,class}, e_t^{h, sem}]\in \mathbb{R}^{D+C+D_{sem}}
\end{equation}
where $e_t^{\text{shared}} \in \mathbb{R}^D$ captures low-level spatio-temporal motion patterns through the shared ViT backbone, $e_t^{h,\text{class}} \in \mathbb{R}^C$ encodes hand-specific classification logits that provide task-aware discriminative signals, and $e_t^{h,\text{sem}} \in \mathbb{R}^{D_{sem}}$ embeds fine-grained semantics into the visual features. 

\subsection{Stage 3: Diffusion-Based Dual-Hand Action Segmentation }
\subsubsection{Architecture}
\textbf{Shared Encoder for Temporal Modeling}. A shared encoder $\mathcal{E}_{seg}$ is employed to process MAS features. $\mathcal{E}_{seg}$ consists of $L_{seg}$ mixed convolution-attention layers:
\begin{equation}
    Z^h, H^h=\mathcal{E}_{seg}(E^h)
\end{equation}
where $Z^h \in \mathbb{R}^{C\times T}$ are initial action logits and $H^h\in \mathbb{R}^{D'\times T}$ are hierarchical backbone features extracted from intermediate encoder layers. 

\textbf{Cross-Hand Feature Fusion}. To address the challenge of complex inter-hand dependencies, we introduce a feature fusion module that enables cross-hand information exchange after hand-specific encoding:
\begin{equation}
H^{LH} = \mathcal{F}^{LH}([H^{LH}; H^{RH}]) + H^{LH}
\end{equation}
\begin{equation}
H^{RH} = \mathcal{F}^{RH}([H^{LH}; H^{RH}]) + H^{RH}
\end{equation}
where $[\cdot; \cdot]$ denotes channel-wise concatenation at each time step and $\mathcal{F}^h$ are hand-specific fusion networks implemented as:
\begin{equation}
\mathcal{F}^h(H) = \text{Conv}_{1\times1}(\text{ReLU}(\text{Conv}_{1\times1}(H)))
\end{equation}
Each fusion network consists of two 1×1 convolutions with ReLU activation, projecting the concatenated $2D'$-dimensional features back to $D'$ dimensions. The residual connection preserves hand-specific information while allowing the model to selectively incorporate cross-hand context. Notably, $\mathcal{F}^{LH}$ and $\mathcal{F}^{RH}$ are separate networks, enabling asymmetric information flow.

\textbf{Forward Process}. Following the improved DDPM \cite{improved_ddpm} and DiffAct \cite{diffact}, we define a forward process that gradually adds Gaussian noise to action distributions. Given ground-truth one-hot encoded action distributions $P^h\in \mathbb{R}^{C\times T}$, it is normalized to range $[-s_{de},s_{de}]$:
\begin{equation}
    X_{0}^h = (2P^h-1)\cdot s_{de}
\end{equation}
The forward process is:
\begin{equation}
    q(X_k^h | X_{0}^h)= \mathcal{N}(\sqrt{\bar{\alpha}_k}X_{0}^h,(1-\bar{\alpha}_k)\mathbf{I})
\end{equation}
where $ \bar{\alpha}_k= \cos^2(\frac{k/K+\delta}{1+\delta}\cdot \frac{\pi}{2})/\cos^2(\frac{\delta}{1+\delta} \cdot \frac{\pi}{2})$.

\textbf{Hand-specific Denoising Decoder}. Each hand-specific decoder predicts clean actions from noisy input $\tilde{P}_k^h$ (denormalized $X_k^h$) conditioned on timestep $k$ and fused features:
\begin{equation}
    \hat{\mathbf{Z}}^h = \mathcal{De}^h(\tilde{\mathbf{P}}_k^h, k, \tilde{\mathbf{H}}^h; \theta^h)
\end{equation}
The decoder uses sinusoidal timestep embeddings and cross-attention to query encoder features. We predict clean actions $\hat{P}_0^h=\mathrm{Softmax}(\hat{Z}^h)$ and derive noise via:
\begin{equation}
    \hat{\epsilon}^h= \frac{X_k^h/\sqrt{\bar{\alpha}_k}-\hat{X}_0^h}{\sqrt{1/\bar{\alpha}_k-1}}
\end{equation}
where $\hat{X}_0^h=\mathrm{clamp}((2\hat{P}_0^h-1)\cdot s_{de},-s_{de},s_{de})$. 

\textbf{DDIM Sampling}. At inference, we use deterministic DDIM with $K'=5$ steps. Starting from $X_{K'}^h\sim \mathcal{N}(\mathbf{0},\mathbf{I})$:
\begin{equation}
    X_{k-1}^h=\sqrt{\bar{\alpha}_{k-1}}\cdot \hat{X}_0^h + \sqrt{1-\bar{\alpha}_{k-1}}\cdot \hat{\epsilon}^h
\end{equation}
\subsubsection{Adaptive Loss Weighting Mechanism}
To balance the training progress on two hands, an adaptive bidirectional loss weighting mechanism is designed to automatically adjust loss weighting based on performance. 

\textbf{Performance Monitoring}. We maintain a sliding window of recent validation accuracies:
\begin{equation}
    \mathcal{W}^h=\{\mathrm{Acc}_i^h\}_{i=ep-w}^{ep}
\end{equation}
where $ep$ is the current epoch and $w$ is the window size.

\textbf{Adaptive Weight Computation}. We compute boost factors based on the performance gaps:
\begin{equation}
\begin{aligned}
    \beta^{LH} = 
\begin{cases} 
\min(\beta_{\max}, \max(\beta_{\min}, \frac{\bar{\mathcal{W}}^{RH}}{\bar{\mathcal{W}}^{LH}})) & \text{if } \frac{\bar{\mathcal{W}}^{LH}}{\bar{\mathcal{W}}^{RH}} < \Delta_{gap} \\
1.0 & \text{otherwise}
\end{cases}
\end{aligned}
\end{equation}
\begin{equation}
\begin{aligned}
    \beta^{RH} = 
\begin{cases} 
\min(\beta_{\max}, \max(\beta_{\min}, \frac{\bar{\mathcal{W}}^{LH}}{\bar{\mathcal{W}}^{RH}})) & \text{if } \frac{\bar{\mathcal{W}}^{RH}}{\bar{\mathcal{W}}^{LH}} < \Delta_{gap} \\
1.0 & \text{otherwise}
\end{cases}
\end{aligned}
\end{equation}
where $\bar{\mathcal{W}}^{h}=\frac{1}{w}\sum_i \mathrm{Acc}_i^h$ is the windowed average, and $[\beta_{min},\beta_{max}]$ are bounds. This bidirectional adjustment automatically increases loss weights for the underperforming hand, preventing performance degradation.

\textbf{Training Loss}. The complete loss function combines encoder and decoder objectives with adaptive weighting:
\begin{equation}
    \begin{aligned}
\mathcal{L}_{\text{total}} &= \sum_{h \in \{LH, RH\}} \beta^h (\lambda_{\text{enc}}^h \mathcal{L}_{\text{enc}}^h + \lambda_{\text{dec}}^h \mathcal{L}_{\text{dec}}^h)
\end{aligned}
\end{equation}
where: 
\begin{equation}
\begin{aligned}
\mathcal{L}_{\text{enc}}^h &= \mathcal{L}_{\text{CE}}(P^h, Y^h) + \lambda_{\text{sm}} \mathcal{L}_{\text{smooth}}(P^h) 
\end{aligned}
\end{equation}
\begin{equation}
\label{quation_loss}
\begin{aligned}
\mathcal{L}_{\text{dec}}^h = \mathcal{L}_{\text{CE}}(\hat{P}_0^h, Y^h) + \lambda_{\text{sm}} \mathcal{L}_{\text{smooth}}(\hat{P}_0^h) +  \\ \lambda_{\text{bd}} \mathcal{L}_{\text{boundary}}(\hat{B}^h, B^h)
\end{aligned}
\end{equation}
Here, $\mathcal{L}_{\text{CE}}$ is cross-entropy with hand-specific class weights computed from training data, $P^h=\mathrm{Softmax}(Z^h)$, $\mathcal{L}_{\text{smooth}}(p_t)=\mathrm{MSE}(\mathrm{log}p_{t+1},\mathrm{log}p_t)$ enforces temporal smoothness. $\mathcal{L}_{\text{boundary}}= \mathrm{BCE}(\hat{B}^h,B^h)$ penalizes boundary errors, where $\hat{B}^h\in [0, 1]^{1\times (T-1)}$ is the predicted boundary and $B^h$ is the ground truth. The loss weights $\lambda_{enc}^h$, $\lambda_{dec}^h$, $\lambda_{sm}$ and $\lambda_{bd}$ are scaled by $\beta^h$ during training.

\section{Experiments}
\subsection{Experimental Setup}
\textbf{Datasets}. We evaluate Polyphony on three datasets covering dual-hand and single-stream scenarios: (1) \textbf{HA-ViD} \cite{ha-vid} is a human assembly video dataset annotated with dual-hand actions, comprising 609 untrimmed videos from three viewpoints (side, front, top), with 75 action classes per hand. (2) \textbf{ATTACH} \cite{attach} is a two-handed assembly dataset containing 378 untrimmed videos across three views, with 24 action classes per hand.  These two datasets are used to assess our method's performance in dual-hand action segmentation—the primary focus of our work. (3) \textbf{Breakfast} \cite{breakfast} is a kitchen activity dataset with 1,712 untrimmed videos, covering 48 action classes. Breakfast is chosen to evaluate the generalization of our approach to conventional single-stream action segmentation in broader domains. For all datasets, we adopt the official splits.

\textbf{Evaluation Metrics}. Following standard protocols in temporal action segmentation, we report frame-wise accuracy (Acc), segmental edit distance (Edit), and segmental F1-scores (F1@{10,25,50}) \cite{ms-tcn}. For dual-hand action segmentation (HA-ViD, ATTACH), we report metrics separately for LH and RH.

\textbf{Implementation Details}. In Stage 1, we sample all segments plus 30, 20, and 5 random clips from each video for HA-ViD (per hand), ATTACH (per hand), and Breakfast, respectively. The visual backbone is a VideoMAE V2 ViT-Base model \cite{videomaev2} pretrained on Kinetics-400 \cite{kinetics400}. ADH-ViT is trained for 50 epochs with an alternation period of $\Delta = 50$ steps, using an AdamW optimizer with a learning rate of 1e-3, cosine annealing scheduler, and weight decay of 0.1. In Stage 2, MiniLM \cite{MiniLM} is employed to extract semantic embeddings. The temporal modeling component consists of three TCN blocks with channel dimensions [512, 128, 64] and kernel size 3. This stage is trained for 100 epochs with an AdamW optimizer (learning rate 3e-4). The adaptive alignment loss is weighted by $\alpha = 0.7$. In Stage 3, the encoder has 10 layers and 64 channels. Each decoder contains 3 cross-attention layers. The diffusion process is configured with $K=1000$ steps. $\lambda_{sm}=0.05$ and $\lambda_{bd}=0.2$ in Equation \ref{quation_loss}. We optimize using Adam with a learning rate of 1e-3 over 1000 epochs. $w=5$, $\Delta_{gap}=0.95$, and $[\beta_{min},\beta_{max}]=[1,2]$ in the adaptive weighting. Three stages are trained sequentially.

\subsection{Main Results}
Polyphony is evaluated against state-of-the-art approaches across the three datasets on both dual-hand and single-stream action segmentation scenarios.

\subsubsection{Dual-hand Action Segmentation}
Tables~\ref{main_havid} and~\ref{main_attach} present results on HA-ViD and ATTACH. With I3D features, the same input as baseline methods, our approach achieves the best performance on most metrics, demonstrating the effectiveness of our diffusion-based segmentation model. With our MAS features, performance gains become substantial: on HA-ViD, ours surpasses the previous best method (FACT) by 12.0 and 16.8 points in left- and right-hand accuracy; on ATTACH, ours outperforms DiffAct by 5.3 and 4.8 percentage points in left- and right-hand accuracy. These improvements extend across all metrics, with Edit distance and F1 scores showing consistent gains of 2-10 points depending on the dataset. As our method performs dual-hand segmentation simultaneously through a single unified model with one shared visual backbone, it is more accurate and appropriate for dual-hand action segmentation scenarios compared with baseline approaches which train separate models for each hand.
\begin{table}[t]
\centering
\captionsetup{skip=4pt}
\caption{Results on HA-ViD Dataset (average over three views, select metrics). Full results in Supplementary Material.}
\resizebox{\linewidth}{!}{%
\begin{tabular}{cccccllcccll}
\hline
\multirow{2}{*}{Method} & \multirow{2}{*}{Input} & \multicolumn{5}{c}{LH}                                               & \multicolumn{5}{c}{RH}                                               \\ \cline{3-12} 
                        &                        & Acc            & Edit           & \multicolumn{3}{c}{F1@10}          & Acc            & Edit           & \multicolumn{3}{c}{F1@10}          \\ \hline
MS-TCN \cite{ms-tcn}& I3D                    & 40.2           & 37.5           & \multicolumn{3}{c}{36.6}           & 39.3           & 34.8           & \multicolumn{3}{c}{34.7}           \\
DTGRM \cite{dtgrm}                   & I3D                    & 40.2           & 37.5           & \multicolumn{3}{c}{39.1}           & 39.7           & 37.3           & \multicolumn{3}{c}{37.8}           \\
BCN \cite{BCN}                     & I3D                    & 44.1           & 41.4           & \multicolumn{3}{c}{43.7}           & 43.4           & 38.0           & \multicolumn{3}{c}{41.3}           \\
C2F-TCN \cite{c2f-tcn}                 & I3D                    & 39.5           & 22.0           & \multicolumn{3}{c}{22.6}           & 39.0           & 21.9           & \multicolumn{3}{c}{22.5}           \\
DiffAct \cite{diffact}                 & I3D                    & 43.7& 42.5& \multicolumn{3}{c}{44.1}          & 44.4& 47.9& \multicolumn{3}{c}{48.8}          \\
FACT \cite{FACT}                    & I3D                    & 45.1& 47.2& \multicolumn{3}{c}{50.1}          & 43.8& 45.3& \multicolumn{3}{c}{47.6}          \\ \hline
\multirow{2}{*}{Ours}   & I3D                    &                45.4&                46.7& \multicolumn{3}{c}{49.9}               &                45.2&                46.7& \multicolumn{3}{c}{47.8}               \\
                        & MAS& \textbf{57.1}& \textbf{53.7}& \multicolumn{3}{c}{\textbf{58.2}} & \textbf{60.6}& \textbf{54.8}& \multicolumn{3}{c}{\textbf{61.5}} \\ \hline
\end{tabular}
}
\vspace{2pt}
{\fontsize{9}{10}\selectfont  
\label{main_havid}}
\end{table}
\begin{table}[]
\centering
\captionsetup{skip=4pt}
\caption{Results on ATTACH Dataset (select metrics). Full results in Supplementary Material.}
\resizebox{\linewidth}{!}{%
\label{tab:my-table}
\begin{tabular}{cccccllcccll}
\hline
\multirow{2}{*}{Method} & \multirow{2}{*}{Input} & \multicolumn{5}{c}{LH}                                               & \multicolumn{5}{c}{RH}                                               \\ \cline{3-12} 
                        &                        & Acc            & Edit           & \multicolumn{3}{c}{F1@10}          & Acc            & Edit           & \multicolumn{3}{c}{F1@10}          \\ \hline
MS-TCN \cite{ms-tcn}                 & I3D                    & 43.5& 46.2& \multicolumn{3}{c}{38.8}          & 36.6& 46.7& \multicolumn{3}{c}{37.7}          \\
C2F-TCN \cite{c2f-tcn}                & I3D                    & 46.3& 19.4& \multicolumn{3}{c}{22.2}          & 40.3& 36.3& \multicolumn{3}{c}{35.2}          \\
DiffAct \cite{diffact}                 & I3D                    & 47.5& 44.1& \multicolumn{3}{c}{40.7}          & 42.5& 46.9& \multicolumn{3}{c}{43.5}          \\
FACT \cite{FACT}                   & I3D                    & 45.8           & 46.8           & \multicolumn{3}{c}{39.1}           & 40.1           & 46.4           & \multicolumn{3}{c}{41.0}           \\ \hline
\multirow{2}{*}{Ours}& I3D                    & 47.2& 47.3& \multicolumn{3}{c}{42.4}          & 42.4& 47.1& \multicolumn{3}{c}{43.5}          \\
                        & MAS& \textbf{52.8}& \textbf{47.8}& \multicolumn{3}{c}{\textbf{45.7}} & \textbf{47.3}& \textbf{49.7}& \multicolumn{3}{c}{\textbf{46.9}} \\ \hline
\end{tabular}
}
\vspace{2pt}
{\fontsize{9}{10}\selectfont  
\label{main_attach}}
\end{table}
\subsubsection{Single-Stream Action Segmentation}
Table~\ref{main_breakfast} presents results on the Breakfast dataset. Despite being designed for dual-hand scenarios, Polyphony can adapt to single-stream settings by blocking one stream or duplicating inputs across both. Results here use the duplication strategy (details in Supplementary Material). With I3D features, we achieve 78.6\% accuracy—establishing new state-of-the-art among I3D-based methods and outperforming DiffAct by 2.2 points. With MAS features, our method sets a new overall state-of-the-art at 82.5\% accuracy, surpassing EAST (82.2\%) despite using a substantially smaller backbone (ViT-Base with 86M parameters vs. ViT-Giant with 1B+ parameters). This demonstrates that Polyphony's effectiveness stems from architectural design and semantic conditioning rather than relying on model scale.
\begin{table}[]
\centering
\captionsetup{skip=4pt}
\caption{Results on Breakfast dataset. EAST employs ViT-Giant (1B+ params, 1408-dim) as the backbone, while our method uses ViT-Base (86M params, 768-dim), achieving better performance with significantly fewer parameters.}
\label{main_breakfast}
\resizebox{0.85\linewidth}{!}{%
\begin{tabular}{ccccc}
\hline
Method    & Input & Acc           & Edit          & F1@10, 25, 50             \\ \hline
MS-TCN \cite{ms-tcn}    & I3D   & 66.3          & 61.7          & 52.6, 48.1, 37.9          \\
DTGRM \cite{dtgrm}     & I3D   & 68.3          & 68.9          & 68.7, 61.9, 46.6          \\
BCN  \cite{BCN}     & I3D   & 70.4          & 66.2          & 68.7, 65.5, 55.0          \\
C2F-TCN \cite{c2f-tcn}   & I3D   & 73.5          & 68.2          & 70.1, 66.6, 56.2          \\
ASFormer \cite{ASformer}  & I3D   & 73.5          & 75.0          & 76.0, 70.6, 57.4          \\
LTContext \cite{ltcontext} & I3D   & 74.2          & 77.0          & 77.6, 72.6, 60.1          \\
DiffAct \cite{diffact}   & I3D   & 76.4          & 78.4          & 80.3, 75.9, 64.6          \\
FACT \cite{FACT}     & I3D   & 76.2          & \textbf{79.7} & 81.4, 76.5, 66.2          \\
Ours& I3D   & \textbf{78.6} & 79.0          & \textbf{81.9, 77.8, 68.3} \\ \hline
EAST \cite{east}      & MAEv2 & 82.2& 83.5          & 85.6, 81.5, 71.6          \\
Ours& MAS   & \textbf{82.5}& \textbf{84.3}& \textbf{86.1, 82.2, 73.2}\\ \hline
\end{tabular}
}
\end{table}

\subsection{Ablation Studies}
We conduct comprehensive ablation studies on HA-ViD to validate our design.
\subsubsection{Component Ablation}
Table~\ref{tab:component_ablation} analyzes the contribution of each component through progressive ablation. The shared \textbf{motion features} (MF) establish a strong baseline by capturing spatio-temporal patterns through the shared ViT backbone. Adding \textbf{action features} (AF) yields mixed results: while Edit distance improves for LH (+1.6\%) and accuracy increases slightly for RH (+0.3\%), LH accuracy marginally decreases (-0.2\%). This suggests that hand-specific classification signals primarily enhance segment-level sequential accuracy rather than raw frame-level accuracy. Incorporating \textbf{semantic features} (SF) produces substantial gains, particularly for RH (Acc: +2.3\%), validating their contribution. The asymmetric improvement (RH benefits more than LH) may reflect that RH, as the typically dominant hand, performs more diverse fine-grained manipulations that benefit from semantic grounding.

Critically, removing either cross-hand \textbf{feature fusion} (FF) or \textbf{adaptive weighting} (AW) not only degrades absolute performance but also exacerbates inter-hand performance imbalance. Without FF, the accuracy gap widens from 3.5\% to 4.0\%, with LH suffering disproportionate degradation (-1.6\% vs. -1.1\% for RH). Removing AW further widens the gap to 4.3\%, with LH plummeting 1.8\% compared to only 1.0\% for RH. This asymmetric impact reveals that without inter-hand modeling and dynamic loss balancing, the dominant hand (RH) monopolizes gradient updates, leaving the non-dominant hand (LH) undertrained. These results validate that both components are essential not only for absolute performance but also for maintaining balanced learning across both hands.

\begin{table}[]
\captionsetup{skip=4pt}
\caption{Ablation on key components. We report the average performance across three views of HA-ViD 
\label{tab:component_ablation}
dataset.}
\resizebox{\linewidth}{!}{%
\begin{tabular}{llllccc}
\hline
 & FF & AW & Hand & Acc & Edit & F1@10, 25, 50 \\ \hline
\multirow{2}{*}{MF} & \multirow{2}{*}{$\checkmark$} & \multirow{2}{*}{$\checkmark$} & LH & 56.0 & 52.7 & 58.6, 52.3, 39.3 \\
 &  &  & RH & 58.0 & 53.0 & 59.7, 53.1, 40.3 \\ \hline
\multirow{2}{*}{MF+AF} & \multirow{2}{*}{$\checkmark$} & \multirow{2}{*}{$\checkmark$} & LH & 55.8 & \textbf{54.3} & \textbf{59.6}, 52.9, \textbf{39.7} \\
 &  &  & RH & 58.3 & 53.4 & 59.1, 52.7, 39.8 \\ \hline
\multirow{2}{*}{MF+AF+SF} & \multirow{2}{*}{$\checkmark$} & \multirow{2}{*}{$\checkmark$} & LH & \textbf{57.1} & 53.7 & 58.8, \textbf{53.0}, 39.6 \\
 &  &  & RH & \textbf{60.6} & \textbf{54.8} & \textbf{61.5}, \textbf{55.1}, \textbf{41.3} \\ \hline
\multirow{2}{*}{MF+AF+SF} & \multirow{2}{*}{$\times$} & \multirow{2}{*}{$\checkmark$} & LH & 55.5 & 53.5 & 57.9, 51.9, 38.6 \\
 &  &  & RH & 59.5 & 54.6 & 60.3, 53.4, 39.3 \\ \hline
\multirow{2}{*}{MF+AF+SF} & \multirow{2}{*}{$\times$} & \multirow{2}{*}{$\times$} & LH & 55.3 & 53.0 & 58.0, 51.9, 38.7 \\
 &  &  & RH & 59.6 & 53.9 & 60.2, 53.9, 40.8 \\ \hline
\end{tabular}
}
\end{table}

\subsubsection{ADH-ViT Study}
\textbf{ADH-ViT as a Standalone Action Recognition Model}. To demonstrate the versatility of our approach, we evaluate ADH-ViT as a standalone action recognition model on the HA-ViD action recognition dataset \cite{ha-vid} (Table~\ref{tab:action_recognition}). While all baseline methods employ separate models for each hand, our ADH-ViT recognizes both hands' actions simultaneously through a single unified model. ADH-ViT (both) achieves competitive performance on both hands with the highest top-5 accuracy on LH (90.4\%). Comparing ADH-ViT (seg) with (both) validates our dual sampling strategy, as incorporating random clip sampling improves top-1 accuracy by 13.1 and 15.2 percentage points for LH and RH respectively. While top-1 accuracy falls below the best baselines, this gap is offset by the reduced parameters and practicality in dual-hand action recognition tasks.

\begin{table}[]
\centering
\captionsetup{skip=4pt}
\caption{Results on HA-ViD action recognition dataset. All baselines use separate models per hand; ours uses one unified model. Results report average top-1/top-5 accuracy across three views. (seg) and (both) denote the use of segmentation-based only and both sampling strategies, respectively.}
\label{tab:action_recognition}
\resizebox{0.80\linewidth}{!}{%
\begin{tabular}{ccccc}
\hline
\multirow{2}{*}{Method} & \multicolumn{2}{c}{LH} & \multicolumn{2}{c}{RH} \\ \cline{2-5} 
                        & Top-1      & Top-5     & Top-1      & Top-5     \\ \hline
TSM \cite{tsm}                     & 61.0       & 88.5      & 58.6       & 87.9      \\
TimeSformer \cite{timesformer}             & 52.1       & 85.4      & 51.8       & 84.4      \\
I3D \cite{i3d}                     & 47.7       & 71.5      & 52.9       & 85.1      \\
MViTv2 \cite{mvitv2}                  & 61.5       & 86.3      & 58.7       & 84.1      \\
UniFormerV2 \cite{uniformerv2}             & \textbf{62.4}& 89.7& 61.4       & \textbf{89.9}\\
VideoMAE V2 \cite{videomaev2}             & 62.1       & 89.0      & \textbf{62.6}& 88.3      \\ \hline
ADH-ViT (seg)           & 47.0       & 75.6      & 46.2       & 74.3      \\
ADH-ViT (both)          &            60.1&           \textbf{90.4}&            61.4&           88.9\\ \hline
\end{tabular}
    }
\end{table}

\textbf{Alternating Training vs. Joint Training}. To validate the efficacy of our alternating training strategy, it is compared against joint training (naive multi-task learning) on both action recognition and segmentation tasks. Table~\ref{tab:training_strategy} demonstrates alternating training's advantages across both dimensions. Joint training leads to a consistent performance gap favoring the right hand. Alternating training improves overall performance and reverses this imbalance in both tasks. Notably, LH benefits more from alternating training (+2.5\% recognition, +3.9\% segmentation) compared to RH (+1.0\%, +0.3\%), indicating that our mechanism specifically addresses the gradient domination problem where the dominant hand monopolizes learning in joint training. 
\begin{table}[]
\captionsetup{skip=4pt}
\caption{Comparison of joint vs. alternating training strategies on action recognition and segmentation tasks. Results on HA-ViD side-view using segment-based sampling for ADH-ViT training.}
\label{tab:training_strategy}
\resizebox{\linewidth}{!}{%
\begin{tabular}{ccccccc}
\hline
\multirow{2}{*}{Training}& \multirow{2}{*}{Hand} & \multicolumn{2}{c}{Action Recognition} & \multicolumn{3}{c}{Action Segmentation} \\ \cline{3-7} 
                                   &                       & Top-1              & Top-5             & Acc      & Edit     & F1@10\\ \hline
\multirow{2}{*}{Joint}             & LH                    &                    44.5&                   75.4&          53.0&          52.9&                   58.6\\
                                   & RH                    &                    45.2&                   \textbf{74.9}&          55.4&          52.7&                   56.0\\ \hline
\multirow{2}{*}{Alternating}       & LH                    &                    \textbf{47.0}       &                   \textbf{75.6}      &          \textbf{56.9}&          \textbf{54.2}&                   \textbf{59.3}\\
                                   & RH                    &                    \textbf{46.2}       &                   74.3      &          \textbf{55.7}&          \textbf{56.0}&                   \textbf{59.3}\\ \hline
\end{tabular}
}
\end{table}

\subsubsection{Semantic Representation Study}
\begin{figure}[h!]
    \centering
    \includegraphics[width=1\linewidth]{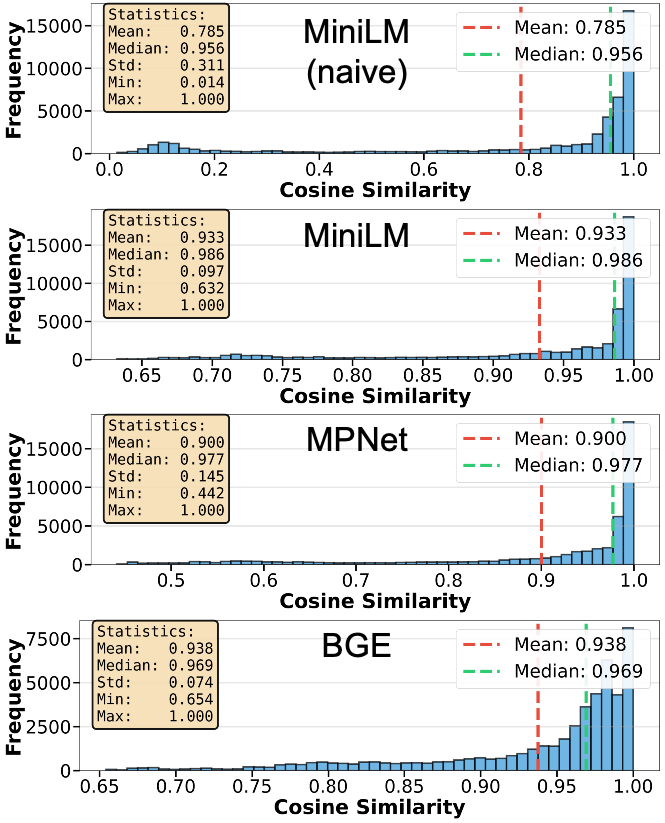}
    \captionsetup{skip=4pt}
    \caption{Distribution of cosine similarities between aligned visual features and semantic embeddings. Higher similarity and tighter distribution indicate better semantic-visual alignment.}
    \label{fig:semantic_embedding}
\end{figure}

While Table~\ref{tab:component_ablation} validates the effectiveness of semantic features, this section provides a deeper analysis of the design choices in our semantic conditioning stage. We investigate three critical aspects: (1) the impact of action description structure (naive vs. structured compositional descriptions), (2) the choice of language models for semantic embedding, and (3) the effectiveness of semantic conditioning on addressing semantic ambiguity challenge. Our analysis combines quantitative performance metrics (Table~\ref{tab:semantic_ablation} and \ref{tab:fine-grained}) with alignment quality assessment (Figure~\ref{fig:semantic_embedding}). Table~\ref{tab:semantic_ablation} demonstrates that structured compositional descriptions substantially outperform naive action descriptions. Figure~\ref{fig:semantic_embedding} shows that structured descriptions produce sharper, more concentrated similarity distributions, indicating stronger semantic-visual alignment. Surprisingly, MiniLM-L6 achieves the best performance, outperforming larger models. This suggests that effective semantic conditioning requires balancing semantic expressiveness with visual feature compatibility—larger models with high-dimensional embeddings may diminish the visual information contribution. We evaluate performance on visually similar but semantically distinct fine-grained actions from HA-ViD side-view (e.g., "screw nut onto bolt" vs. "screw nut onto shaft"). Table~\ref{tab:fine-grained} shows semantic features improve accuracy by 3.6 points on average, validating their role in disambiguation.
\vspace{-2mm}
\begin{table}[]
\centering
\captionsetup{skip=4pt}
\caption{Results on HA-ViD side-view dataset using different language models. Results averaged across both hands.}
\label{tab:semantic_ablation}
\resizebox{\linewidth}{!}{%
\begin{tabular}{ccccccc}\toprule
Model           &Description& Dim  & Size& Acc & Edit & F1@10 \\\midrule
MiniLM-L6 \cite{MiniLM}&Naive& 384  & 22M  &     55.0&      52.9&       57.8\\
MiniLM-L6 \cite{MiniLM}&Structured& 384  & 22M  &     57.4&      53.5&       57.4\\
MPNet-base \cite{mpnet}     &Structured& 768  & 110M &     56.4&      53.4&       58.3\\
BGE-large \cite{bge}      &Structured& 1024 & 335M &     56.6&      53.3&      
58.6\\ \bottomrule\end{tabular}
}
\end{table}
\vspace{-2mm}

\vspace{-5mm}
\begin{table}
    \centering
\captionsetup{skip=4pt}
\caption{Segmentation accuracy on visually similar but semantically distinct fine-grained actions.}
\label{tab:fine-grained}
\resizebox{0.50\linewidth}{!}{%
    \begin{tabular}{ccc}\toprule
         Method&  LH&  RH\\\midrule
         + semantic features&  54.8&  54.9\\
         - semantic features&  48.7&  53.8\\ \hline
         Improvement&  +6.1&  +1.1\\ \bottomrule
    \end{tabular}
    }
\end{table}

\subsubsection{Qualitative Segmentation Results}
\begin{figure}
    \centering
    \includegraphics[width=1\linewidth]{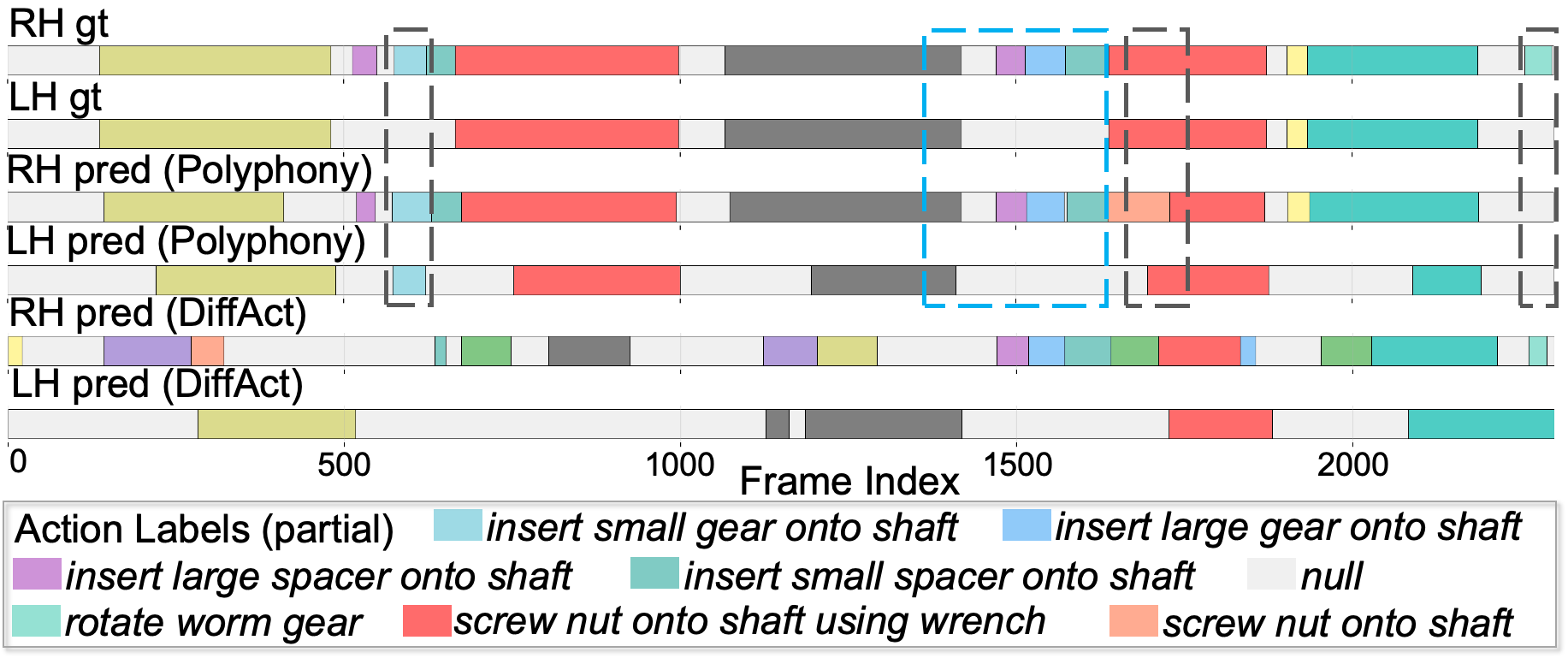}
    \captionsetup{skip=4pt}
    \caption{Qualitative segmentation results on a sample from HA-ViD. Top: Ground truth for RH and LH. Middle: Our predictions. Bottom: DiffAct baseline. The blue box highlights successful segments and gray boxes indicate failure cases for discussion in the text. Selected action labels are shown for discussion.}
    \label{fig:qualitative}
\end{figure}

Figure \ref{fig:qualitative} presents qualitative segmentation results of Polyphony and DiffAct on a HA-ViD video for both hands. Polyphony (middle) produces more coherent predictions with more accurate action transitions and better inter-hand coordination compared to DiffAct (bottom), which exhibits severe over-segmentation and poor coordination. 

The blue box highlights successful cases of our method: (1) correctly predicting asymmetric actions where hands perform different actions simultaneously; (2) distinguishing fine-grained semantic variations; (3) maintaining temporal synchronization when both hands execute coordinated actions. These successes collectively demonstrate that our approach addresses the core challenges of complex inter-hand dependencies, visual asymmetry, and semantic ambiguity in dual-hand action segmentation.

However, the gray boxes reveal failure modes that expose current limitations: (left) over-predicting hand coordination when the ground truth shows independent actions, suggesting our model has learned a coordination bias from training data; (middle) confusing tool-mediated actions and bare-handed actions, indicating a limited visual discrimination capability; (right) missing short, long-tail actions (rotate worm gear). These results reveal challenges with hand independence modeling, visual ambiguity, and long-tail action classes that motivate future work.
\vspace{-2mm}
\vspace{-3mm}

\section{Conclusion}
\vspace{-2mm}
This paper introduced Polyphony, a unified three-stage method that addresses dual-hand action segmentation through an Alternating Dual-Hand Vision Transformer (ADH-ViT), Semantic Feature Conditioning, and Diffusion-based Dual-Hand Action Segmentation. Polyphony achieves state-of-the-art on both dual-hand benchmarks (HA-ViD, ATTACH) and the single-stream Breakfast benchmark, validating its effectiveness across a variety of tasks. The framework's modular and versatile design, where each stage can be independently improved and ADH-ViT can operate as a standalone action recognition model, positions Polyphony as a foundational architecture for future research extending beyond bimanual scenarios to multi-agent collaborative behavior understanding.

While Polyphony achieves strong performance, several limitations remain. Our method exhibits learned coordination biases that over-predict hand synchronization and struggles with visually ambiguous tool-mediated actions and long-tail classes. Its reliance on manual construction of semantic descriptions limits scalability. Future directions include end-to-end approaches, self-supervised learning to reduce annotation dependence, automated semantic description generation via large language models, and extending our framework to multi-agent collaborative scenarios.

\vspace{-3mm}
\section*{Acknowledgements}
\vspace{-2mm}
Work supported by CAIR and CQTS funded by Tamkeen NYUAD RI Award CG010 and CG008, respectively.

{
    \small
    \bibliographystyle{ieeetr}
    \bibliography{main}
}


\end{document}